\documentclass[sigconf, nonacm, 9pt]{acmart}

\usepackage{geometry}
\usepackage[T1]{fontenc}

\usepackage{graphicx}

\usepackage[ruled,vlined]{algorithm2e}
\usepackage{ragged2e}
\usepackage{multirow}
\usepackage{color,soul}
\definecolor{myGreen}{rgb}{0.25,0.60,0.15}
\definecolor{myRed}{rgb}{0.85,0.40,0.30}
\definecolor{myPink}{rgb}{0.85,0.30,0.80}
\newcommand{\green}[1]{{\textbf{\color{myGreen}#1}}} 
\newcommand{\red}[1]{{\color{myRed}#1}}

\newenvironment{usecase}[1][htb]{
    
    \begin{algorithm}[#1]%
    }{
    \end{algorithm}
}

\begin{document}

\title{Embedding Domain-Specific Knowledge from LLMs into the Feature Engineering Pipeline}


\author{João Eduardo Batista}
\email{joao.batista@riken.jp}
\orcid{0000-0002-2997-8550}
\affiliation{%
  \institution{RIKEN-CCS}
  \city{Kobe}
  \country{Japan}
}

\begin{abstract}
Feature engineering is mandatory in the machine learning pipeline to obtain robust models. While evolutionary computation is well-known for its great results both in feature selection and feature construction, its methods are computationally expensive due to the large number of evaluations required to induce the final model. Part of the reason why these algorithms require a large number of evaluations is their lack of domain-specific knowledge, resulting in a lot of random guessing during evolution. In this work, we propose using Large Language Models (LLMs) as an initial feature construction step to add knowledge to the dataset. By doing so, our results show that the evolution can converge faster, saving us computational resources. The proposed approach only provides the names of the features in the dataset and the target objective to the LLM, making it usable even when working with datasets containing private data. 
While consistent improvements to test performance were only observed for one-third of the datasets (CSS, PM, and IM10), possibly due to problems being easily explored by LLMs, this approach only decreased the model performance in 1/77 test cases.
Additionally, this work introduces the M6GP feature engineering algorithm to symbolic regression, showing it can improve the results of the random forest regressor and produce competitive results with its predecessor, M3GP.
\end{abstract}

\keywords{
Genetic Programming, 
Feature Engineering,
Symbolic Regression, 
Large Language Models,
AI4Science
}

\maketitle

\section{Introduction}

\par 
Feature engineering~\cite{CI} is a mandatory step in the machine learning pipeline that allows the users to select features that are relevant to the problem while discarding potentially noisy ones~(feature selection), and transform/combine features into new ones to capture underlying patterns in the data~(feature construction). Although several algorithms are commonly used for automatic feature engineering, such as Principal Component Analysis~(PCA)~\cite{pca}, Fast Feature Extraction~(FFX)~\cite{ffx}, Genetic Algorithms~(GA)~\cite{ga}, and Genetic Programming~(GP)~\cite{poli08}, whenever possible, researchers use their domain-specific knowledge to improve the dataset, biasing the models into learning the data correctly.

\par  
Evolutionary Computation~(EC)~\cite{Eiben2015} is well-known for its capability in feature engineering tasks, with GAs being one of the main options for feature selection. GP algorithms can perform feature construction with implicit feature selection, since the final model combines the original dataset features while potentially ignoring some of them. While these EC algorithms obtain good results in both classification~\cite{Nguyen2024} and regression~\cite{Chen2024} tasks, even in out-of-distribution data~\cite{Batista20,transfer}, it is computationally expensive to induce models due to a lot of random guessing involved in the evolutionary cycle, especially in early generations where the population consists of randomly generated models.

\par 
Considering the recent developments in Large Language Models (LLMs)~\cite{llm,naveed,attention}, LLMs are better at answering questions about any topic. While there are still some issues regarding LLMs not informing the user about hallucinated content~\cite{Li23}, the community agrees that LLMs are typically correct when dealing with questions about well-known topics. So, we start this work with the hypothesis that LLMs can recommend feature combinations in datasets, allowing us to use them to embed knowledge into the datasets in an initial feature engineering step. While we consider it worth the risk to include bad feature combinations (that can be later ignored) in the dataset, we do not perform feature selection with the LLMs, since we risk throwing away good features permanently.

\par 
This work contributes to the GP field in two ways: by extending previous work on M6GP~\cite{m6gp} to symbolic regression applications, showing that we can obtain robust feature engineering models not only for classification datasets, and by proposing a pipeline that can reduce the computation cost required to induce models. It also contributed to the AI4Science community by showing a reliable way to use LLMs to produce robust models without compromising private data.

\section{Related Work}
\par 
Evolutionary computation is well-known for its capabilities in feature engineering, generating models robust to out-of-distribution data~\cite{Batista20,transfer}. In addition to their robustness, EC algorithms also have the potential to induce interpretable models~\cite{ecxai,xaigp,Zhou2024}, which also reflects itself in models that can be applied to weaker machines due to the low computational cost of the induced models. While these qualities make EC a powerful choice for many tasks, the computation cost involved in inducing the final model is often too high for users to consider training the model. This cost tends to increase as the predictive strength of the model goes up. Take SLUG~\cite{ex_slug} for example, the authors themselves recognize that, although the algorithm has great feature selection capabilities, the evolution is ``sluggish.'' Wrapper-based algorithms like M3GP~\cite{m3gp}, M5GP~\cite{m5gp}, and M6GP~\cite{m6gp} adapt their feature engineering capabilities to the wrapped algorithm. To do so, when calculating the fitness of an individual, they induce a model (classifier~\cite{batista22,m6gp,MA2023101285}/regressor~\cite{m5gp,transfer,m3gp_reg}) from an algorithm given as a parameter and use its performance as fitness. Since these algorithms use a population size of 500 and 100 generations by default, one can quickly notice that to induce a feature engineering model, the algorithm has to induce 50.000 other models.

\begin{table*}[t]
    \centering
    \caption{Datasets used in this work, number of features in the original dataset and their names, no. classes and target label, no. samples and no. features including those generated by GPT-4o. The PM/PT, HL/CL, and IM3/IM10 datasets were extracted from the same dataset so, they contain the same original.}
    \setlength{\tabcolsep}{2pt}
    \begin{tabular}{l|p{7.2cm}|p{2.9cm}|c|c}
        Dataset   & No. Original Features and Names  & No. classes and Target & Samples &  Features  \\
        \hline
    \textbf{Regression Datasets}&&&&\\
        Energy Efficiency (HL) & \multirow{2}{7.1cm}{\textbf{(8)} Relative Compactness, surface area, wall area, roof area, overall height, orientation, glazing area, glazing area distribution} & 
        Housing Heating Load & 768 & 12
           \\ 
        Energy Efficiency (CL) & & Housing Cooling Load & 768 & 13 \\
         &&&\\  
        Concrete (CCS)     & \textbf{(8)} Cement, blast furnace slag, fly ash, water, superplasticizer, coarse aggregate, fine aggregate, age & Concrete Compression Strength  & 1030 & 20     \\
        Istanbul Stocks (IS)  & \textbf{(7)} SP, DAX, FTSE, NIKKEI, BOVESPA, EU, EM  & USD-based ISE  & 536 & 9   \\
        Parkinson Motor (PM)  & 
        \multirow{2}{7.0cm}{\textbf{(19)} Age, sex, test time, jitter (\%), jitter abs., jitter RAP, jittep PPQ5, jitter DDP, Shimmer, shimmer dB, shimmer apq3, shimmer apq5, shimmer apq11, shimmer DDA, NHR, HNR, RPDE, DFA and PPE} & Motor UPDRS & 5875 & 32 \\
        Parkinson Total (PT) & &Total UPDRS & 5875 & 32 \\
        &&&&\\ &&&&\\
        \hline
        \textbf{Classification Datasets}&&&&\\
        Heart (HRT)         & \textbf{(13)} Age, sex, chest pain type, resting bp, serum cholesterol, fasting blood sugar, ECG results, max heart rate, angina, old peak, slope of the peak exercise ST segment, number of major vessels colored by fluoroscopy, thal  & \textbf{(2)} Presence of heart disease                                & 270  & 24 \\
        Image-3 (IM3)       & \multirow{2}{7.7cm}{\textbf{(6)} Landsat-7 satellited bands 1--5 and 7}  & \textbf{(3)} Forest types          & 322  & 10 \\
        Image-10 (IM10)     &  & \textbf{(10)} Land cover types     & 6797 & 13 \\
        Yeast (YST)         & \textbf{(8)} MCG, GVH, ALM, MIT, ERL, POX, VAC, and NUC  & \textbf{(6)} Location site                                & 1484 & 14 \\
        Student Academic Success (SAS) &  \textbf{(36)} Marital status, application mode, application order, course, daytime/evening attendance, previous qualification, grade of previous qualification, nationality, mothers and fathers qualification and occupation, admission grade, displaces, special needs, debtor, tuition fees up to date, gender, scholarship holder, age at enrollment, is international, number of units credited, enrolled evaluated approved graded and without evaluation in the first and second semester, unemployment rate, inflation rate, GDP & \textbf{(3)} Dropout, enroll or graduate, at the end of the normal duration of the course&4424& 43 \\
    \end{tabular}
    \label{datasets}
\end{table*}


\par 
LLMs are being used to extract features from tabular data by providing them as input~\cite{llmfe,fsllm,tablet}. In this work, we are using LLMs without providing them the dataset samples, providing only the names of the features and our target goal, and relying only on the LLM's internal knowledge and retrieval-augmented generation~(RAG)~\cite{rag2,rag} capabilities, as will be explained in the next section. While we consider this to be the best use of LLMs in terms of dealing with private data (although all our datasets are public), we acknowledge that by providing the datasets to the LLM, we can obtain better results by generating better features and even perform data augmentation~\cite{augtab,llmir}, something not feasible in our pipeline. In addition to the privacy issue, processing data and generating features can be a ``premium'' feature (i.e., free users have limited access to these features), as seen in ChatGPT~\cite{chatgpt}, making our approach accessible to all users.

\par 
Currently, there is an increasing interest in applying LLMs to the industry and all scientific fields, with EC being no exception. In a work by Jorgensen~\cite{giorgia} \textit{et al.}, LLMs are used to increase the number of test cases when training GP agents, resulting in better agents. Lehman \textit{et al.}~\cite{Lehman2024} use LLMs as a mutation operator for code generation by prompting the LLM to generate mutations to the individuals based on the target domain, increasing the likelihood of the mutation resulting in better offspring. Meyerson \textit{et al.}~\cite{Meyerson24} propose an LLM-based crossover operator. Similarly, Jin \textit{et al.}~\cite{eot} uses LLMs as an evolutionary optimizer, letting LLMs combine and mutate prompts during evolution. Guo \textit{et al.}~\cite{evoprompt} also uses evolutionary algorithms to optimize LLM prompts.

\section{Methodology}

\subsection{Datasets}

\par 
We use six regression datasets and five classification datasets, obtained from the UCI repository~\cite{UCI} (all datasets except IM3 and IM10) and from the U.S. Geological Survey (USGS) Earth Resources Observation Systems Data Centre~\cite{EROS} (IM3 and IM10). The dataset names, features, number of samples, and objectives are displayed in Table~\ref{datasets}. As will be later explained, we use GPT-4o~\cite{gpt4} as a feature construction method, allowing us to extend the number of features of the HL, CL, CCS, IS, PM, and PT datasets by 4, 5, 12, 2, 13, and 13 features, respectively, in the regression datasets. In the classification datasets, the number of features was extended by 8, 3, 4, 6, and 7, in the HRT, IM3, IM10, YST, and SAS datasets, respectively. Readers familiar with previous work on M3GP and M6GP may notice that four datasets of the typical classification benchmark were not used. This decision was based on the lack of documentation on the meaning of each variable (MOVL and WAV datasets), the objective of the dataset being unfit for this work (SEG), and the dataset not being found at UCI (VOW).

\begin{figure*}[t]
    \centering
    \includegraphics[width=0.8\linewidth]{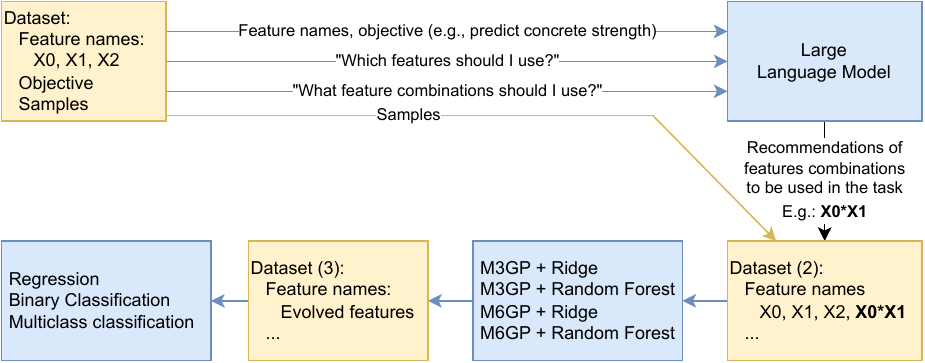}
    \caption{Proposed pipeline with two feature engineering steps. 
    An LLM recommends feature combinations based on the available features and objectives, without accessing the dataset. These combinations are then added to the dataset, and M3GP or M6GP is used as a second feature engineering step.}
    \label{pipeline}
\end{figure*}

\subsection{Algorithms}
We use a total of 6 components in our experimental setups: GPT-4o, accessed through its web app~\cite{chatgpt}, to embed domain-specific knowledge into the datasets; M3GP and M6GP, using their \textit{m3gp} and \textit{m6gp} Python libraries~\cite{batista22,m6gp}, for feature engineering; and the Ridge, Decision Tree~(DT), Random Forest~(RF) regressors, using the implementations available at the $sklearn$ Python library~\cite{scikit}, to learn and make predictions on the dataset. The parameters of the M3GP and M6GP algorithms are seen in Table~\ref{parameters}.

\begin{table}[t]
    \centering
    \setlength{\tabcolsep}{4pt}
    \caption{Parameters used in our experimental setup.}
    \label{parameters}
    \begin{tabular}{ll}
         \textbf{General parameters}                & \textbf{Value}  \\
         \hline
         \hspace{1mm}Number of trials               & \hspace{1mm} 30 \\
         \hspace{1mm}Training and test size         & \hspace{1mm} 70\% and 30\% of the dataset \\         
         \hspace{1mm}$p$-value                      & \hspace{1mm} 0.01   \\
                  \\
         \textbf{DT/RF parameters}                  & \textbf{Value}  \\
         \hline
         \hspace{1mm} Max depth                     & \hspace{1mm} 6   \\
         \hspace{1mm} No. estimators (RF)           & \hspace{1mm} 100    \\
\\
         \textbf{GP parameters}                     & \textbf{Value}  \\
         \hline
         \hspace{1mm}Population size                & \hspace{1mm} 500   \\
         \hspace{1mm}No. generations                & \hspace{1mm} 100 (Ridge) and 30 (RF)    \\
         \hspace{1mm}Initialization                 & \hspace{1mm} 6-depth Full initialization   \\
         \hspace{1mm}Operator probabilities:        &  \\
         \hspace{1mm}-Crossover operators           & \hspace{1mm} Both operators: 1/4    \\
         \hspace{1mm}-Mutation operator             & \hspace{1mm} All 3 operators: 1/6    \\
         \hspace{1mm}Function set                   & \hspace{1mm} +, -, x, // protected division~\cite{m3gp} \\
         \hspace{1mm}Terminal set                   & \hspace{1mm} Dataset features   \\
         \hspace{1mm}Bloat control                  & \hspace{1mm} 17-depth limit  \\
         \hspace{1mm}Wrapped algorithm              & \hspace{1mm} Ridge, or random forest\\
         \textbf{M3GP}&\\
         \hspace{1mm}Fitness (regression)           & \hspace{1mm} 2-fold RMSE (untied with size) \\
         \hspace{1mm}Fitness (classification)       & \hspace{1mm} 2-fold WAF (untied with size) \\
         \hspace{1mm}Selection                      & \hspace{1mm} Single tournament \\
         \hspace{1mm}Elitism                        & \hspace{1mm} Best individual of the generation \\
         \textbf{M6GP}&\\
         \hspace{1mm}Fitness (multi-objective)      & \hspace{1mm} 2-fold RMSE, and MAE (regression) \\
         \hspace{1mm}Fitness (multi-objective)      & \hspace{1mm} 2-fold WAF, and Size (classification) \\
         \hspace{1mm}Selection                      & \hspace{1mm} Double tournament~\cite{m6gp} \\
         \hspace{1mm}Elitism                        & \hspace{1mm} First non-dominated front 
    \end{tabular}
\end{table}

\par \textbf{GPT-4o~(GPT)~\cite{chatgpt}:} We use the ChatGPT web app to obtain recommendations for feature combinations from the GPT-4o LLM. LLMs are trained on vast amounts of data from different fields. Additionally, through RAG, they can retrieve information in real-time and add it to the query, enriching the answer. Thanks to this knowledge, they generate answers on virtually any topic. However, there is a risk of the answer being a hallucination, i.e., incorrect~\cite{xie24}. Additionally, LLMs are typically stochastic, reducing the reproducibility of this part of the pipeline;

\par \textbf{M3GP~\cite{m3gp}:} M3GP induces multi-tree models that convert the input dataset into a new dataset, optimized to the fitness function. Each tree in the model is an arithmetic expression that combines dataset features, making M3GP a feature engineering algorithm. The fitness function selected as a parameter (2FOLD-RMSE and 2FOLD-WAF) splits the output dataset into two halves, induces a model in each half, and calculates their Root Mean Squared Error~(RMSE) or Weighted Average F-score (WAF) in the other half. By using the average RMSE/WAF of the two models as fitness (2-fold cross-validation), we reduce overfitting~\cite{batista22};

\par \textbf{M6GP~\cite{m6gp}:} M6GP is a direct successor of M3GP. While M3GP is a single-objective algorithm, M6GP allows for multiple objectives. In this case, we optimize performance (2-fold cross-validation, like the M3GP), the maximum absolute error (regression datasets), and size (classification datasets), reducing the likelihood of obtaining predictions with a large error in regression datasets;

\par \textbf{Ridge~\cite{rid}:} Ridge is a linear regression algorithm that includes L2 regularization and uses the linear least squares function as the loss function. We picked this algorithm for regression due to its simplicity, aiming to verify its competitiveness with algorithms with more predictive power, such as the RF;

\par \textbf{Decision Tree (DT)~\cite{dt}:} DT is a supervised algorithm that infers decision rules, being applicable to both regression and classification problems. We use a maximum depth of 6, limiting the model size to 64 parameters;

\par \textbf{Random Forest (RF)~\cite{rf}:} RF is a supervised algorithm that induces an ensemble of DT models, being used for both classification and regression problems as well. We use RF models with 100 estimators (DT models) with a maximum depth of 6, limiting the model size to 6400 parameters. Due to previous work suggesting that 30 generations are enough for M3GP to learn feature engineering models fit for RF~\cite{batista22}, we limit M3GP and M6GP to 30 generations when using RF in the fitness function.

\subsection{Feature Engineering Pipeline}

We propose a pipeline~(see Figure~\ref{pipeline}) that consists of two feature engineering steps before the final model is induced: first, we perform feature construction using an LLM without considering the data itself, relying only on the labels and the objective of the dataset, e.g., ``predicting concrete strength''; the second step converts a dataset into a new one, optimizing for a specific task (e.g., symbolic regression) and model (Ridge or RF) using GP-based feature engineering:

\par 
\textbf{Domain-specific knowledge:} We ask an LLM~(GPT-4o) for recommendations on which dataset features are relevant to the task, filtering features that are not related to the objective. Then, we ask which combinations of those features may improve our results. Use cases~\ref{llm_example} and~\ref{llm_concrete} illustrate the template we used for the prompts of each dataset and its application to the CSS dataset, respectively.
 
\begin{usecase}[t]
\caption{Generic use case of feature engineering using an LLM.} 
\par \justifying{ \hspace{-5mm} \textbf{USER:} I want to [objective] in a dataset with the following features: [feature names]. Which features should I use?}
\par \justifying{ \hspace{-5mm} \textbf{LLM:} (Provides a sub-set of the dataset features related to that objective.)}
\par \justifying{ \hspace{-5mm} \textbf{USER:} Are there any combination of features that might improve the results?}
\par \justifying{ \hspace{-5mm} \textbf{LLM:} (Gives a list of feature combinations related to that objective, using the previously selected features.)}
 \label{llm_example}
\end{usecase}
\begin{usecase}[t]
\caption{Feature engineering on the CSS dataset using GPT-4o.} 
\par \justifying{ \hspace{-5mm} \textbf{USER:} I want to predict concrete compressive strength in a dataset with the following features: cement, Blast furnace slag, fly ash, water, superplasticizer, coarse aggregate, fine aggregate, age. Which features should I use?}
\par \justifying{ \hspace{-5mm} \textbf{LLM:} (GPT-4o recommends using all features, explaining the meaning of each one. The answer is too long to display in the paper.)}
\par \justifying{ \hspace{-5mm} \textbf{USER:} Are there any combination of features that might improve the results?}
\par \justifying{ \hspace{-5mm} \textbf{LLM:} (GPT-4o makes 12 recommendations, such as ratios between features, summing features within a category (e.g., cementitious materials, or aggregate materials), and using log or polynomial transformations on the age feature to reflect a nonlinear evolution of the concrete strength over time. The answer is too long to display in the paper.)}
 \label{llm_concrete}
\end{usecase}

\par 
LLMs can hallucinate, resulting in wrong answers. While the feature combinations recommended by LLMs may be noisy, they can be later thrown away in the second feature engineering step. As such, the benefits of obtaining useful combinations compensate for the risk of obtaining bad ones. To avoid permanently removing useful features, we do not perform feature selection using the LLM. Since GPT-4o recommends features without seeing dataset samples, only their labels, there is no risk of data contamination. As such, this task is only performed once, and the same set of features is used in all 30 runs.

\par
\textbf{Feature engineering with GP:} After adding the new feature combinations to the dataset, we proceed with the pipeline proposed in the reference papers~\cite{batista22,m6gp}. The modified dataset is split into training and test sets, and the training set is used to obtain a feature engineering model using other algorithms, such as M3GP~\cite{m3gp} and M6GP~\cite{m6gp}. Both algorithms use multi-tree models that convert the input dataset into a new dataset where each feature is the output of each tree in the model. To evaluate each model, a model (Ridge or RF) is induced in this new dataset, and its performance is used as fitness, as previously explained, optimizing the feature engineering to the wrapped algorithm.

\section{Results and Discussion}
\par
This section is split into discussing the results from three points of view. First, we compare the results obtained when using GPT for feature construction, or not, allowing us to verify the advantages of using GPT-based feature construction. Then, we compare the results of all 14 experiments: DT, RF, Ridge, M3GP, and M6GP wrapped in the Ridge models (M3GP-Ridge and M6GP-Ridge), and M3GP and M6GP wrapped in the RF models (M3GP-RF and M6GP-RF), with and without GPT-based feature construction, further validating the M3GPs capability for feature engineering, and showing M6GP can also be used for symbolic regression. Lastly, we discuss the size and dimensionality of the GP-based feature engineering models, commenting on the computational costs of using the induced models.

\begin{table*}[t]
\centering
\caption{Median test RMSE in each experiment. $p$-values smaller than 0.01 are highlighted in green/red, indicating significantly better/worse results when using GPT-4o for feature construction, respectively.}
\begin{tabular}{l|cccccc|ccccc}
&\multicolumn{6}{c|}{Regression (RMSE)}&\multicolumn{5}{c}{Classification (WAF)}\\
Datasets&HL&CL&CSS&IS&PM&PT& HRT & IM3 & IM10 & YST & SAS \\
\hline
\textbf{GPT-based Feature Engineering} &&&&\\
\hline
DT&0.583&2.023&8.050&0.018&4.987&5.906&0.759&0.938&0.898&0.578&0.756\\
DT-GPT&0.578&1.995&7.164&0.018&4.986&5.987&0.737&0.949&0.902&0.565&0.763\\
\quad$p$-value&0.976&0.929&\green{0.000}&0.451&0.894&0.280&0.773&\green{0.007}&\green{0.001}&0.117&0.019\\
\hline
RF&0.578&1.781&6.195&0.015&4.255&4.997&0.827&0.959&0.907&0.616&0.791\\
RF-GPT&0.590&1.840&5.744&0.015&4.158&4.997&0.829&0.964&0.921&0.614&0.790\\
\quad$p$-value&0.399&0.058&\green{0.000}&0.918&\green{0.000}&0.988&0.482&0.070&\green{0.000}&0.894&0.767\\
\hline
Ridge&3.057&3.274&10.432&0.019&7.549&9.775&0.859&0.823&0.699&0.582&0.791\\
Ridge-GPT&3.051&3.186&6.901&0.016&7.429&9.697&0.841&0.824&0.758&0.583&0.793\\
\quad$p$-value&0.745&0.114&\green{0.000}&\green{0.000}&\green{0.000}&0.014&0.162&0.712&\green{0.000}&0.756&0.179\\
\hline
\textbf{GP+GPT-based Feature Engineering} &&&&\\
\hline
M3GP-Ridge&0.500&1.523&6.146&0.043&6.954&9.140&0.817&0.927&0.880&0.594&0.784\\
M3GP-Ridge-GPT&0.501&1.611&5.897&0.024&6.963&9.193&0.827&0.933&0.878&0.594&0.784\\
\quad$p$-value&0.723&0.026&\green{0.004}&0.124&0.110&0.139&0.813&0.261&0.243&0.906&0.918\\
\hline
M6GP-Ridge&0.692&1.787&6.715&0.017&7.466&9.821&0.815&0.897&0.761&0.585&0.795\\
M6GP-Ridge-GPT&0.619&1.829&6.377&0.017&7.451&9.795&0.814&0.883&0.808&0.590&0.801\\
\quad$p$-value&0.344&0.657&\green{0.000}&0.701&0.231&0.894&0.641&0.098&\green{0.000}&0.344&0.135\\
\hline
M3GP-RF&0.501&1.379&5.607&0.015&2.192&3.165&0.827&0.969&0.934&0.606&0.782\\
M3GP-RF-GPT&0.513&1.372&5.524&0.015&2.252&3.229&0.816&0.959&0.931&0.608&0.780\\
\quad$p$-value&0.416&0.894&0.322&0.564&0.110&0.025&0.784&0.387&0.101&0.636&0.679\\
\hline
M6GP-RF&0.513&1.458&5.841&0.015&2.592&3.541&0.805&0.959&0.916&0.612&0.769\\
M6GP-RF-GPT&0.527&1.492&5.618&0.015&2.660&3.780&0.805&0.938&0.920&0.607&0.777\\
\quad$p$-value&0.249&0.790&0.107&0.965&0.352&\red{0.001}&0.506&0.399&0.016&0.605&0.030\\
\hline
\end{tabular}
\label{results_table}
\end{table*}

\subsection{Feature Construction using GPT-4o}

\par
Table~\ref{results_table} shows the median test RMSE (regression) and WAF (classification) obtained over 30 runs in all experiments, with and without feature engineering. In the non-GP experiments, we apply the Ridge, DT, and RF models to the datasets and, as expected, RFs obtain the best results in all datasets, except for SAS (tied with Ridge) and HRT (lower WAF than Ridge, using a $p$-value of 0.01). However, we should consider that these models use 100 estimators and a maximum depth of 6, resulting in models with a size of up to 6400 parameters. By contrast, the DTs have up to 64 parameters (maximum depth 6), and Ridge has one parameter for each feature in the dataset, plus one. This gives an advantage to RFs in terms of predictive power, while making them more computationally expensive to use. This table also shows the comparison of results between the baseline (not using GPT features) and using GPT for feature construction. Interestingly, there is no statistically significant impact in 5 out of 11 datasets. However, this approach brings consistent significant improvements in the CSS, PM, and IM10 datasets, and in one experiment in the IS and IM3 datasets, and significantly significant WAF reduction in one experiment in the PT dataset. Since the CSS and IM10 tasks use features that have well-known meanings, this seems to imply that GPT is better at recommending features for well-studied problems.

\par
\begin{figure*}[t]
    \centering
    \includegraphics[width=0.379\linewidth, trim={0 35 0 17}, clip]{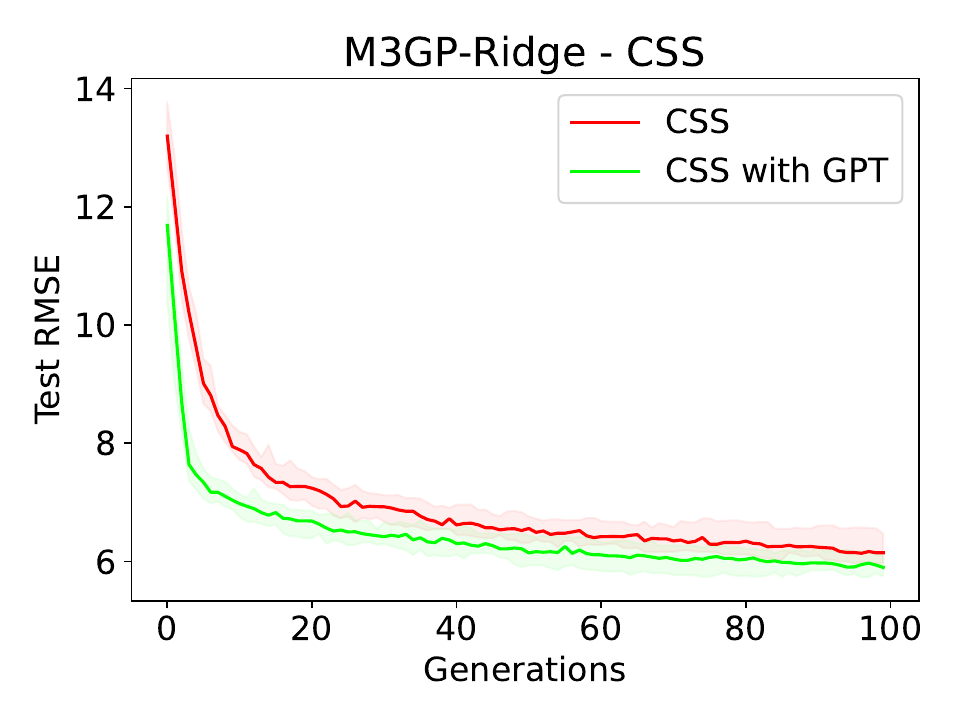}
    \includegraphics[width=0.379\linewidth, trim={0 35 0 17}, clip]{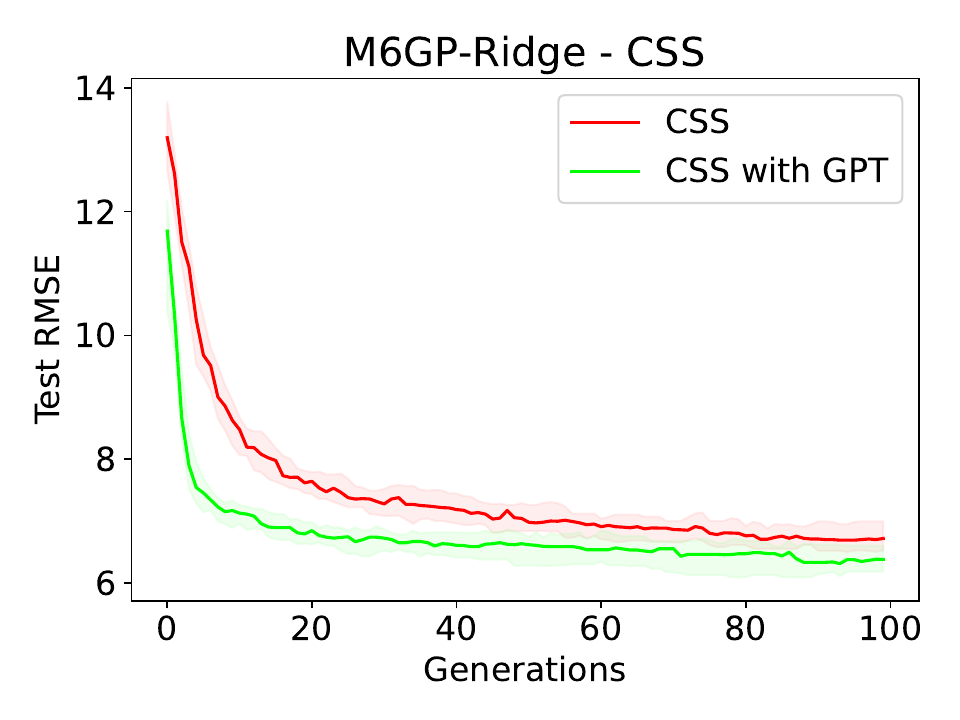}\\
    \includegraphics[width=0.379\linewidth, trim={0 17 0 17}, clip]{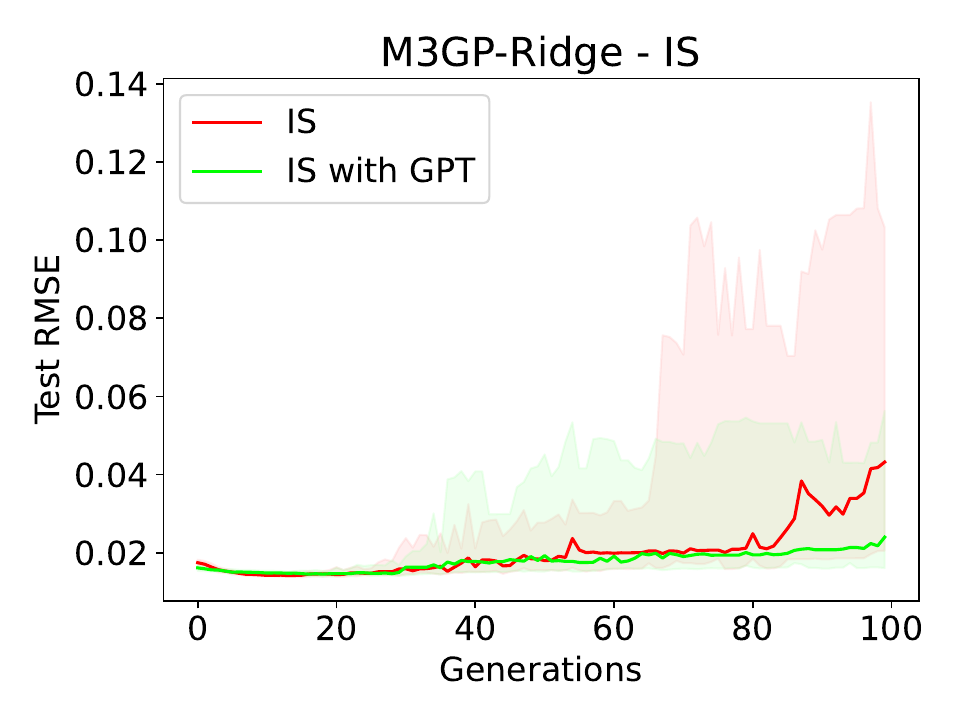}    
    \includegraphics[width=0.379\linewidth, trim={0 17 0 17}, clip]{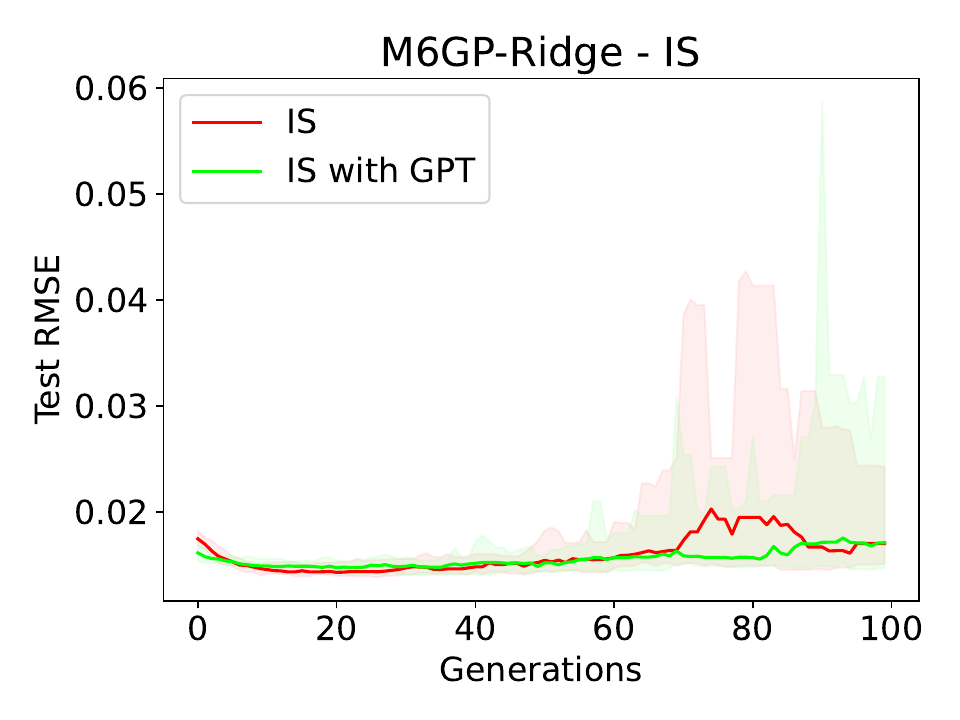}    
    \caption{Median test RMSE over 100 generations when using M3GP-Ridge and M6GP-Ridge in the CSS and IS datasets. The plots highlight the space between quartiles 1 and 3, showing a large dispersion of values when inducing models using IS.}
    \label{mrid_css}
\end{figure*}

 
One of the goals of this work is to show that adding domain-specific knowledge through GPT can accelerate the learning rate of evolutionary computation algorithms. Take the M3GP-Ridge and M6GP-Ridge experiments on the CSS into consideration. In Figure~\ref{mrid_css}, we see the evolution of the test RMSE over the generations and notice that, as expected from the $p$-value, there is a clear advantage in using GPT. Using the median RMSE from Ridge-GPT (6.901) as a baseline, we see that it took 33/13 and 64/15 generations for M3GP-Ridge/M3GP-Ridge-GPT and M6GP-Ridge/M6GP-Ridge-GPT to surpass it, respectively.

\begin{figure*}[t]
    \centering
    \includegraphics[width=0.379\linewidth, trim={0 17 0 17}, clip]{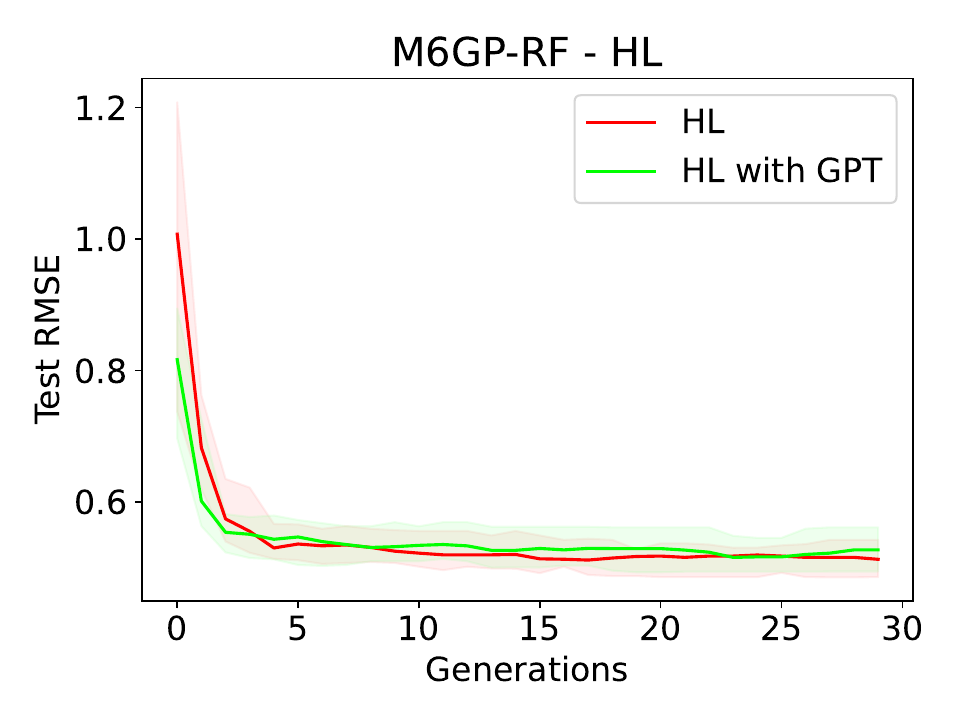}
    \includegraphics[width=0.379\linewidth, trim={0 17 0 17}, clip]{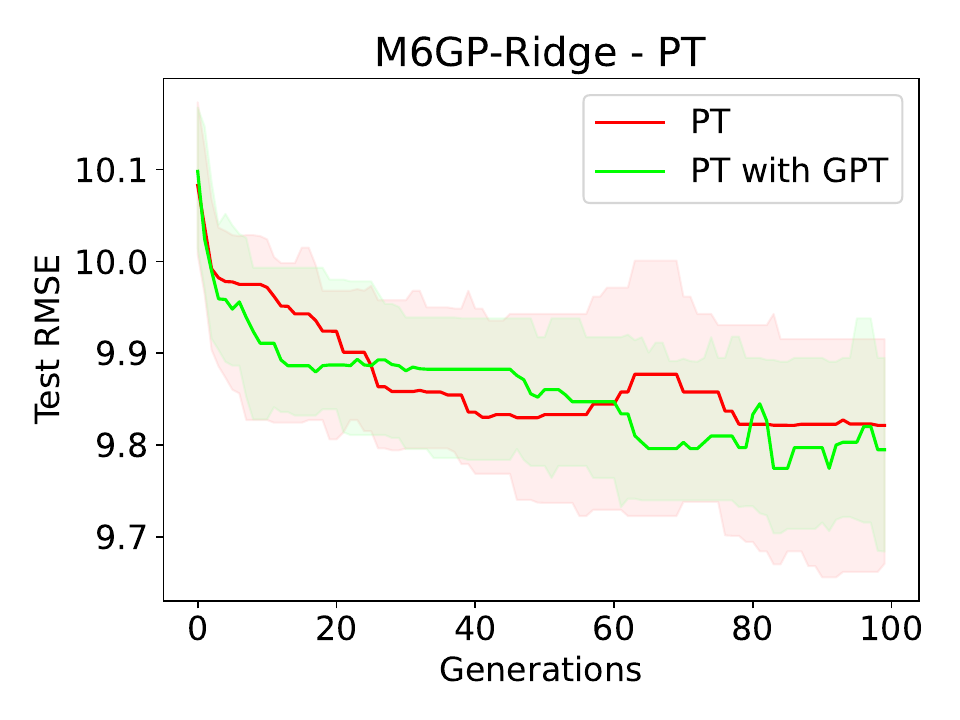}\\
    \caption{Median test RMSE over 30 generations when using M6GP-RF and M6GP-Ridge in the HL and PT datasets, respectively. The plots highlight the space between quartiles 1 and 3. These plots are representative of the other datasets. }
    \label{mr6_css}
\end{figure*}

\begin{figure*}[t]
    \centering
    \includegraphics[width=0.379\linewidth, trim={0 17 0 17}, clip]{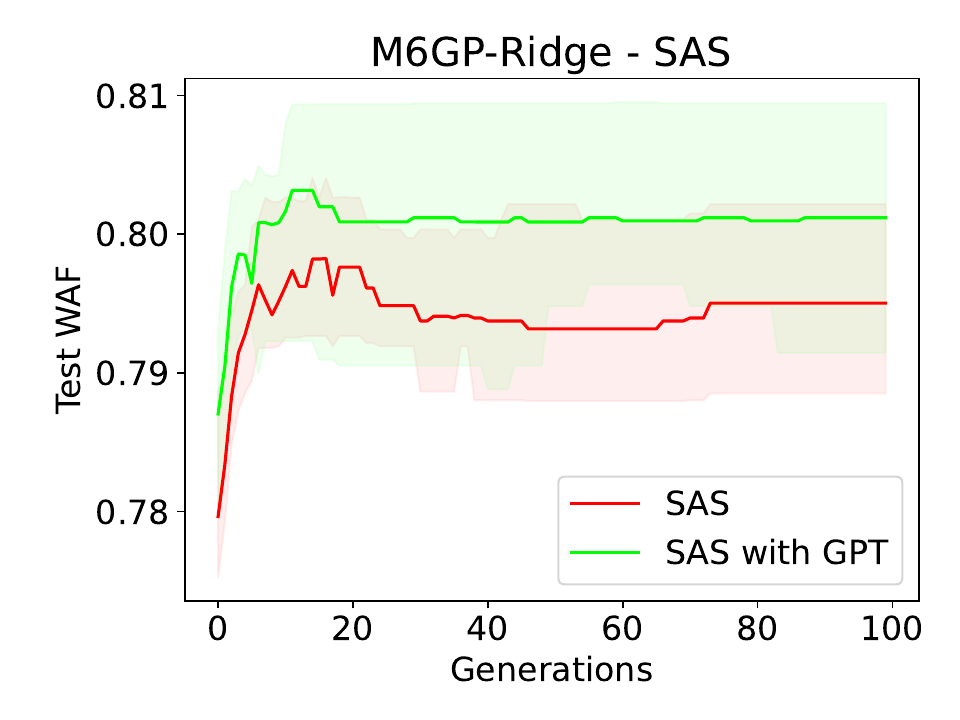}
    \includegraphics[width=0.379\linewidth, trim={0 17 0 17}, clip]{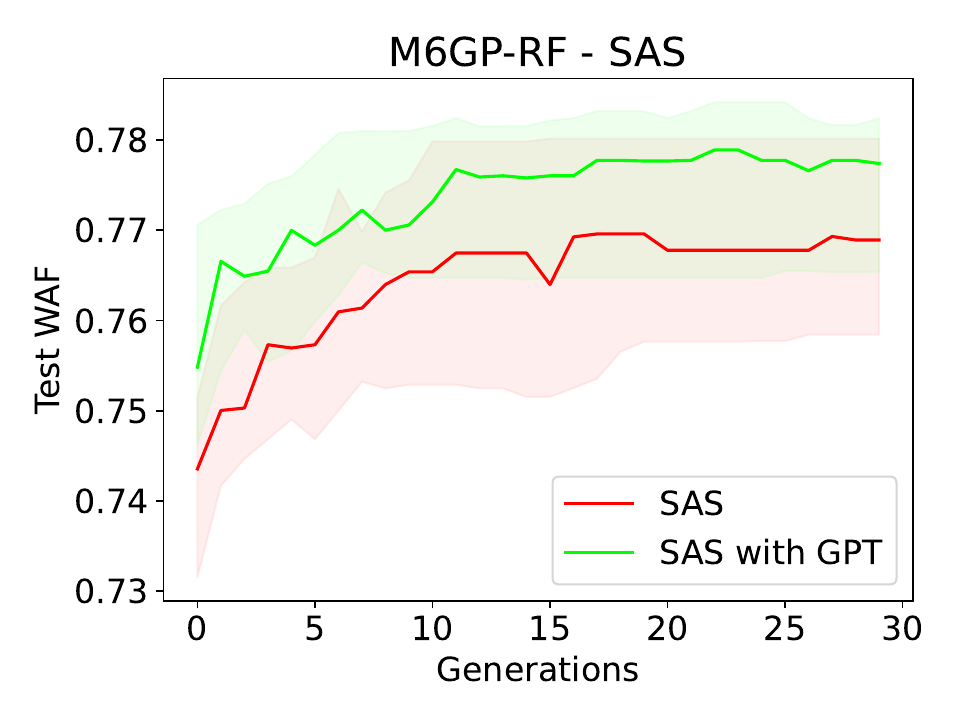}\\
    \caption{Median test WAF over 30 generations when using M6GP-Ridge and M6GP-RF in the SAS dataset. The plots highlight the space between quartiles 1 and 3. }
    \label{m6_class}
\end{figure*}

\par
While we do not see significant improvements in the M3GP-Ridge and M6GP-Ridge experiments in the IS dataset, the plots suggest a decrease in overfitting by using GPT. While both setups obtain their best performance in early generations and start to overfit over time, there is a negative shift in the performance of the M3GP-Ridge models after generation 60, which led to the median test RMSE doubling from generation 84 to generation 100. When using M6GP-Ridge, we see something similar. And, while the baseline recovers from the overfitting spike, the results in the last generation are equivalent to when using GPT features.

\par 
Figure~\ref{mr6_css} shows the evolution of the test RMSE when using M6GP-RF in the HL and PM datasets. In these datasets, and the others (the plots are not shown in the paper), we see that the inclusion of GPT features in the dataset brings little to no change in the evolutionary progress. This further indicates that, while in some cases these features may improve the results or possibly decrease overfitting (CSS and IS datasets), GP is robust to bad feature combinations. Given that this approach takes little time to implement and only needs to be done once per dataset, we consider that this approach is showing promising results.

\par 
Figure~\ref{m6_class} shows the evolution of the test WAF when using M6GP-Ridge and M6GP-RF in the SAS dataset. In the classification experiments using M3GP, we did not see any substantial improvements in the number of generations required to induce a model when using GPT features. However, M6GP shows improvements when using these features. We hypothesize that, since M6GP models for classification tend to be very small, adding good (according to GPT) feature combinations to the dataset compensates for the ``size" objective in the fitness function. With the GPT features, the model can use a combination of features at the cost of increasing its size by 1.

\par 
From a prompt engineering point of view, the query used was very simple (see Use Case~\ref{llm_example}) and general enough to apply to all datasets. We tried two other approaches that we consider even more direct: asking directly for combinations of features in a single prompt, and asking for useful features and combinations, also in a single prompt. We noticed that when using the first approach, the features included in the combinations were sometimes not related to the problem, i.e., the feature selection step was sometimes skipped. So, we decided to try the second approach and noticed that, while GPT was able to select relevant features consistently, the feature construction step was not consistent enough for us to rely on it. By separating the second prompt into two, asking first for relevant features and then for feature combinations, we obtained approximately the same set of combination recommendations across different GPT sessions, making this step of the pipeline less stochastic.

\begin{table*}[t]
\centering
\caption{Median test RMSE in each experiment. This table is identical to Table~\ref{results_table}, except that here $p$-values higher than 0.01 are highlighted in green, indicating no significant difference between that experiment and the one with the best median.}
\setlength{\tabcolsep}{8pt}
\begin{tabular}{l|cccccc|ccccc}
&\multicolumn{6}{c|}{Regression (RMSE)}&\multicolumn{5}{c}{Classification (WAF)}\\
Datasets&HL&CL&CSS&IS&PM&PT& HRT & IM3 & IM10 & YST & SAS \\
\hline
DT&0.583&2.023&8.050&0.018&4.987&5.906&0.759&0.938&0.898&0.578&0.756\\
DT-GPT&0.578&1.995&7.164&0.018&4.986&5.987&0.737&0.949&0.902&0.565&0.763\\
\hline
RF&0.578&1.781&6.195&\green{0.015}&4.255&4.997&\green{0.827}&\green{0.959}&0.907&\green{0.616}&0.791\\
RF-GPT&0.590&1.840&5.744&\green{0.015}&4.158&4.997&0.829&\green{0.964}&0.921&\green{0.614}&0.790\\
\hline
Ridge&3.057&3.274&10.432&0.019&7.549&9.775&\green{0.859}&0.823&0.699&0.582&0.791\\
Ridge-GPT&3.051&3.186&6.901&0.016&7.429&9.697&\green{0.841}&0.824&0.758&0.583&0.793\\
\hline
M3GP-Ridge&\green{0.500}&1.523&6.146&0.043&6.954&9.140&0.817&0.927&0.880&0.594&0.784\\
M3GP-Ridge-GPT&\green{0.501}&1.611&5.897&0.024&6.963&9.193&0.827&0.933&0.878&0.594&0.784\\
\hline
M6GP-Ridge&0.692&1.787&6.715&0.017&7.466&9.821&0.815&0.897&0.761&0.585&\green{0.795}\\
M6GP-Ridge-GPT&0.619&1.829&6.377&0.017&7.451&9.795&0.814&0.883&0.808&0.590&\green{0.801}\\
\hline
M3GP-RF&\green{0.501}&\green{1.379}&\green{5.607}&\green{0.015}&\green{2.192}&\green{3.165}&0.827&\green{0.969}&\green{0.934}&\green{0.606}&0.782\\
M3GP-RF-GPT&\green{0.513}&\green{1.372}&\green{5.524}&\green{0.015}&\green{2.252}&3.229&0.816&\green{0.959}&\green{0.931}&\green{0.608}&0.780\\
\hline
M6GP-RF&\green{0.513}&\green{1.458}&5.841&\green{0.015}&2.592&3.541&0.805&\green{0.959}&0.916&\green{0.612}&0.769\\
M6GP-RF-GPT&0.527&1.492&\green{5.618}&\green{0.015}&2.660&3.780&0.805&0.938&0.920&\green{0.607}&0.777\\
\hline
\end{tabular}
\label{results_rank}
\end{table*}

\subsection{Overall Results}
\par
Take Table~\ref{results_rank} into consideration. Here, we see the experimental setup with the lowest median RMSE and highest median WAF highlighted in green. The results with no statistically significant difference from the best setup are also highlighted. This indicates that, e.g., five setups obtained the best results in the HL dataset, and that M3GP-RF obtained the best results in all datasets (although sometimes tied).

\par 
While we previously stated that the RF had an advantage over the other methods thanks to their size alone, we see that the M3GP-Ridge approach can outperform RFs (HL and CL datasets) and M6GP-Ridge outperformed all other methods in the SAS dataset, while using a model that has less than one-tenth of the size (as we will later see in Table~\ref{results_dim}). Since both M3GP-Ridge and M6GP-Ridge are optimizing features for Ridge, they improved Ridge's results in most datasets. While this improvement is less noticeable in the CSS dataset and, when using M6GP-Ridge, in the PM and PT datasets, the usefulness of these feature engineering methods becomes very obvious in the CL dataset (reducing the median RMSE to half) and in the HL dataset (reducing the median RMSE to one fifth). Additionally, while there was no clear optimal method for the SAS dataset before using GP-based feature engineering, M6GP-Ridge and M6GP-Ridge-GPT obtained the best results. 
By wrapping M3GP and M6GP around a model with a stronger predictive power (RF), we see that we obtain statistically significant improvements in 5 out of 6 regression datasets, and tying with RF and RF-GPT in the IS dataset. And, in 1 out of 5 classification datasets, with results tied in 4 datasets (M3GP-RF). While M6GP obtains similar results (either statistically equivalent or within a 2\% WAF drop), the model sizes and much lower that M3GP models by up to a 95\% reduction (as will be later seen in Table~\ref{results_dim}).

\par 
Overall, these results further reinforce that M3GP is capable of performing feature engineering for symbolic regression, as seen in previous works~\cite{m3gp_reg,transfer}, while showing that M6GP is also capable of obtaining statistically similar results in most datasets, while still outperforming the non-GP experimental setups. M6GP is a new algorithm that has not been sufficiently explored yet. These results further motivate exploring its feature engineering capabilities, potentially improving these results in the future.

\subsection{Model Size and Dimensionality}

In Table~\ref{results_dim}, we see the median model size (i.e., the number of nodes, or parameters) within the M3GP and M6GP models in each experiment, as well as their median dimensionality (number of evolved features/trees within the model). Interestingly, adding the feature combinations suggested by GPT does not have a statistically significant impact on the model dimensionality in most of the experiments, only having an impact on the M3GP-RF models induced using the IS dataset. 

\begin{table*}[t]
\centering
\caption{Median model dimensionality in each experiment. The red values indicate statistically significant growth using $p$-value of 0.01. In most cases, GPT-4o does not affect the size/dimensionality of the GP models.}
\setlength{\tabcolsep}{6.5pt}
\begin{tabular}{l|cccccc|ccccc}
&\multicolumn{6}{c|}{Regression (RMSE)}&\multicolumn{5}{c}{Classification (WAF)}\\
Datasets&HL&CL&CSS&IS&PM&PT& HRT & IM3 & IM10 & YST & SAS \\
No. features (with GPT)& 19 (22)& 19 (22)& 8 (20)& 7 (9)&  8 (21)& 8 (21) & 13 (24) & 6 (10) & 6 (13) & 8 (14) & 36 (43) \\
\hline
Model Size&&&&&&\\
\quad M3GP-Ridge&526.0&493.0&679.0&739.5&630.5&599.0&191.0&154.0&711.5&438.5&220.0\\
\quad M3GP-Ridge-GPT&591.5&496.0&662.5&723.5&581.0&559.5&218.0&167.0&463.5&420.5&342.5\\
\hline
\quad M6GP-Ridge&506.0&384.0&445.0&503.5&447.0&391.0&7.0&24.5&19.0&16.0&5.0\\
\quad M6GP-Ridge-GPT&486.0&364.0&444.0&415.5&371.5&408.0&6.0&26.0&13.5&15.5&6.0\\
\hline
\quad M3GP-RF&97.5&113.0&198.5&203.5&89.0&86.5&140.5&62.5&212.0&163.5&129.0\\
\quad M3GP-RF-GPT&73.0&80.0&150.5&\red{348.0}&92.0&107.0&162.0&90.0&99.5&212.5&93.0\\
\hline
\quad M6GP-RF&64.5&99.5&154.0&136.5&86.0&94.0&8.0&8.0&9.0&7.0&6.0\\
\quad M6GP-RF-GPT&80.5&136.5&135.0&132.0&123.5&103.5&6.0&9.0&8.0&8.0&5.0\\
\hline
Model Dimensionality&&&&&&\\
\quad M3GP-Ridge&21.0&19.0&25.0&16.0&23.0&24.0&8.0&6.0&24.0&18.0&19.0\\
\quad M3GP-Ridge-GPT&22.0&20.0&24.0&15.0&24.0&25.0&10.0&7.0&24.0&18.5&19.0\\
\hline
\quad M6GP-Ridge&24.0&22.0&22.5&16.5&17.5&14.0&3.0&4.0&6.0&7.0&3.0\\
\quad M6GP-Ridge-GPT&24.0&20.0&22.0&15.0&17.0&14.5&4.0&2.0&6.0&7.0&4.0\\
\hline
\quad M3GP-RF&6.0&6.0&9.0&6.0&7.0&6.0&5.5&5.0&11.0&10.0&8.5\\
\quad M3GP-RF-GPT&6.5&5.5&10.0&\red{10.0}&7.0&7.0&6.0&4.0&10.5&10.0&9.0\\
\hline
\quad M6GP-RF&5.0&7.0&9.0&7.5&6.0&7.0&2.0&3.0&5.0&6.0&4.0\\
\quad M6GP-RF-GPT&5.0&7.0&9.0&7.0&6.5&7.0&2.0&3.0&6.0&6.0&3.0\\
\hline
\end{tabular}
\label{results_dim}
\end{table*}


\par 
In the regression part of the table, while there is some variation in the model size across different experiments, their dimensionality is mostly stable, depending exclusively on the dataset and the wrapped algorithm. Since Ridge performs a linear combination of the dataset features, it requires a higher number of features to avoid underfitting to the data, explaining why this number is much larger than that seen in the RF experiments. In the classification part of the table, it is impossible to compare fairly the size of M3GP and M6GP models since the later aims to minimize size, resulting in much smaller models in all datasets, in both size and dimensionality. Despite this size difference, M3GP-RF and M6GP-RF were tied among the best results in the IM3 and YST datasets.

\par 
One important thing to note from this take is that, considering the computational cost of the induced models, M3GP-Ridge and M6GP-Ridge may be good alternatives to using the RF approach without GP-based feature engineering, in regression tasks. As previously seen, GP-based feature engineering allows Ridge to obtain competitive results with the RF models in 4 out of 6 regression datasets. However, M3GP-Ridge and M6GP-Ridge have a total model size ranging from approximately 370 to 760 learned parameters. Meanwhile, the RF models used can have up to 6400 parameters, resulting in models that are more computationally expensive to use, and potentially less interpretable.


\section{Conclusions}

This work proposes a two-step feature engineering pipeline, where the first step uses a Large Language Model~(LLM) to embed domain-specific knowledge into the dataset and the second step uses Genetic Programming~(GP) to perform both feature selection and construction. We apply this pipeline to both regression and classification datasets, but it can be easily adapted to other tasks. By including an LLM in the pipeline, the dataset provided as input to the GP algorithm contains recommendations on which features may work together to solve the dataset. Thanks to this, the initial population contains more useful information, in contrast to simply containing randomly generated individuals. However, the LLM does not access the data directly, relying on the user to input the names of the features and a description of the dataset objective. Thanks to this, this pipeline prevents private data from being leaked into the LLM web application. The results indicate that, while the LLM does not have access to the dataset samples, it can still consistently improve the test performance of the models in one-third of the datasets, while only reducing the test performance in 1/77 test cases. The results indicate that this approach works best for well-studied problems.

\par 
Additionally, this work extends previous work on the M3GP and M6GP. While these two feature engineering algorithms are typically applied to classification tasks, M3GP has already shown good results when dealing with feature engineering for symbolic regression tasks~\cite{transfer,m3gp_reg}. M6GP, being a very recent algorithm, has no previous work on this topic. The results show that M3GP and M6GP can improve the results of well-known methods, such as random forests, in symbolic regression tasks. While the features provided by GPT resulted in improvements in the accuracy of the ridge and random forest regressors with little effort, the results show that the best experimental setup tried is using GP-based feature engineering.

\par 
While GPT-based feature construction produced good results, we intend to keep exploring this approach in the future, exploring other approaches, such as reasoning models, to obtain feature recommendations. M6GP~\cite{m6gp} was initially proposed as a multi-objective wrapper-based feature engineering algorithm that could prevent the bloat that exists in the M3GP models by including structural complexity metrics as an objective. In this work, we could not find complexity metrics that could be used without ruining the performance of the models, an issue also mentioned by the authors. The complexity function used by the authors for classification is not usable in symbolic regression tasks so, future work also includes obtaining structural complexity functions that can be used by M6GP for symbolic regression tasks.


\bibliographystyle{splncs04}
\bibliography{main}

\end{document}